# Machine Learning Algorithms to Predict Chess960 Result & Develop Opening Themes

Shreyan Deo
DPS Vasant Kunj, Delhi, India

Nishchal Dwivedi
Department of Basic Science and Humanities, SVKM's NMIMS Mukesh Patel School of Technology Management & Engineering, Mumbai, India

**Abstract**

This work focuses on the analysis of Chess 960, also known as Fischer Random Chess, a variant of traditional chess where the starting positions of the pieces are randomized. The study aims to predict the game outcome using machine learning techniques and develop an opening theme for each starting position. The first part of the analysis utilizes machine learning models to predict the game result based on certain moves in each position. The methodology involves segregating raw data from .pgn files into usable formats and creating datasets comprising approximately 500 games for each starting position. Three machine learning algorithms- KNN Clustering, Random Forest, and Gradient Boosted Trees- have been used to predict the game outcome. To establish an opening theme, the board is divided into five regions: center, white kingside, white queenside, black kingside, and black queenside. The data from games played by top engines in all 960 positions is used to track the movement of pieces in the opening. By analysing the change in the number of pieces in each region at specific moves, the report predicts the region towards which the game is developing. These models provide valuable insights into predicting game outcomes and understanding the opening theme in Chess 960.

Keywords: Chess 960, Fischer Random Chess, machine learning, game outcome prediction, opening theme, KNN Clustering, Neural Networks, Gradient Boosted Trees.

Note: Trying to see just by looking at the snapshot of an evolved game how accurately we can predict who will win

## 1. Introduction

Chess 960, also known as Fischer Random Chess, is a variant of the traditional chess game. It was invented by the former World Chess Champion, Robert Fischer, to introduce more creativity and reduce the impact of opening theory in the game. Chess 960 is played on the same board as regular chess, but the starting positions of the pieces are randomised, providing a different setup for each game. (Gligoric, 2003) The first part of this report uses machine learning to try and predict the game's outcome. The machine learning model uses data from certain moves of each position and tries to predict the result. The importance of Chess 960 lies in its ability to challenge players in new and exciting ways. Traditional chess has a vast opening theory (Sterren, 2009); players often spend significant time memorising and analysing different opening moves. This can sometimes lead to a reliance on memorised lines rather than genuine creative thinking.

Chess 960 breaks away from this dependency on memorisation. With randomised starting positions, players must rely on understanding chess principles, logical thinking, and strategic planning. This levels the playing field and allows for more level-headed competition. However, there is still a need for a primary starting point for a player, and this report tries to establish that. In the second part of this report, the data of the various games played by top engines in all 960 positions is used to outline how the game progresses in the opening. For this purpose, the board has been divided into five roughly equal regions:

*Table 1: Regions of the Chess Board with chess notation*

| S.No | Region | No. of Squares | Squares |
|---|---|---|---|
| 1 | Centre | 12 | c4, c5, d3, d4, d5, d6, e3, e4, e5, e6, f4, f5 |
| 2 | White Kingside | 13 | h1, h2, h3, h4, g1, g2, g3, g4, f1, f2, f3, e1, e2 |
| 3 | White Queenside | 13 | a1, a2, a3, a4, b1, b2, b3, b4, c1, c2, c3, d1, d2 |
| 4 | Black Kingside | 13 | h8, h7, h6, h5, g8, g7, g6, g5, f8, f7, f6, e8, e7 |
| 5 | Black Queenside | 13 | a8, a7, a6, a5, b8, b7, b6, b5, c8, c7, c6, d8, d7 |

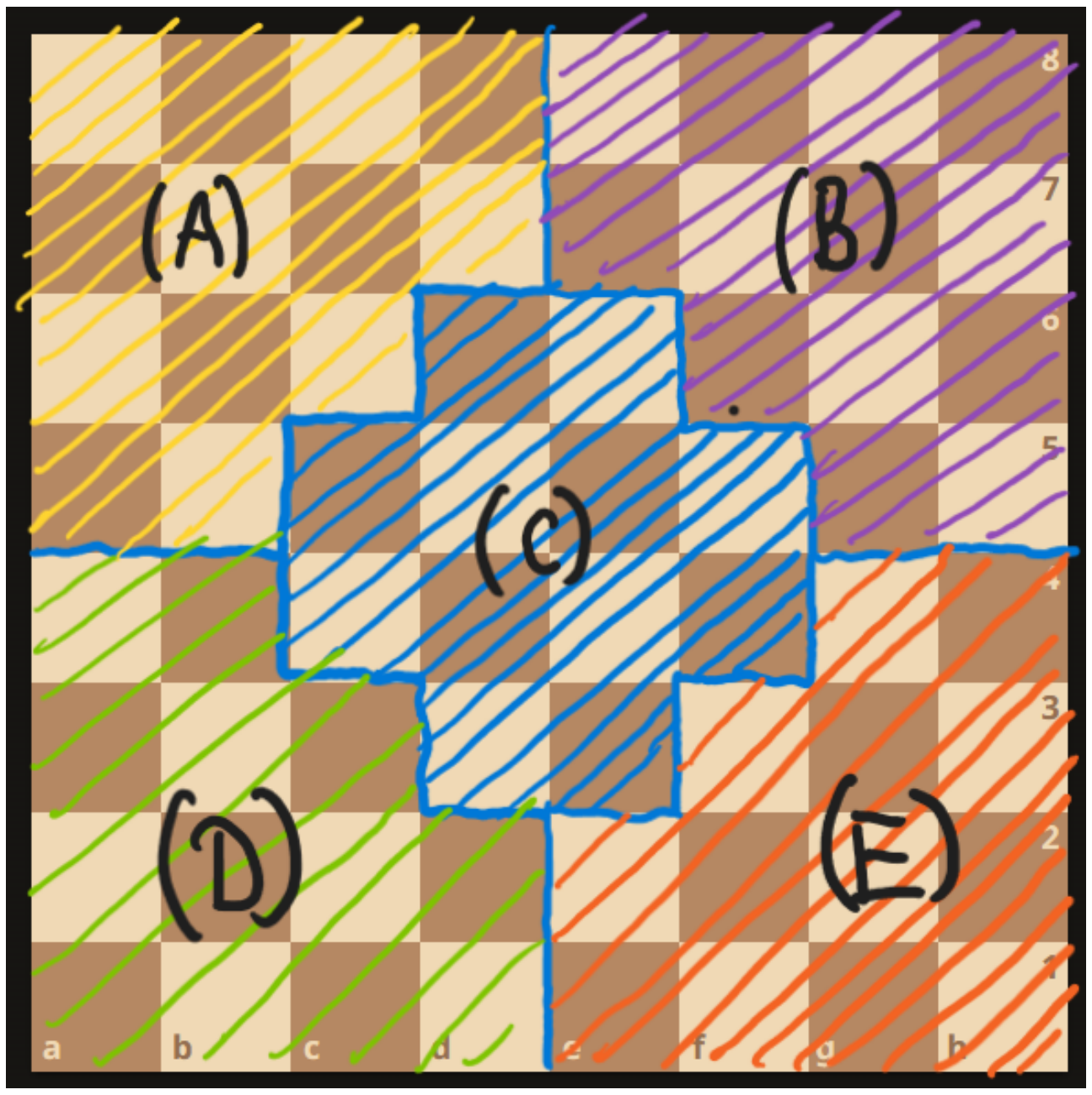

*Figure 1: Regions of the Chess Board*
Region A = Black Queenside
Region B = Black Kingside
Region C = Centre

Region D = White Queenside
Region E = White Kingside

Specifically, this report maps the transition of the game in the opening: to which of the regions the pieces tend to move in a specific starting position. Based on this, all the starting positions are grouped into categories.

## 2. Methodology

### 2.1. Segregation of Data

All the raw data available in the form of .pgn files (which contain games in the standard chess notation) was systematically put into a usable format. This involved segregating the data by putting the required data in tables for faster functioning. We used 500 games for each of the 960 starting positions. Our data set comprised of 480,000 games in total.

### 2.2. Visualising Chess Positons as Numbers

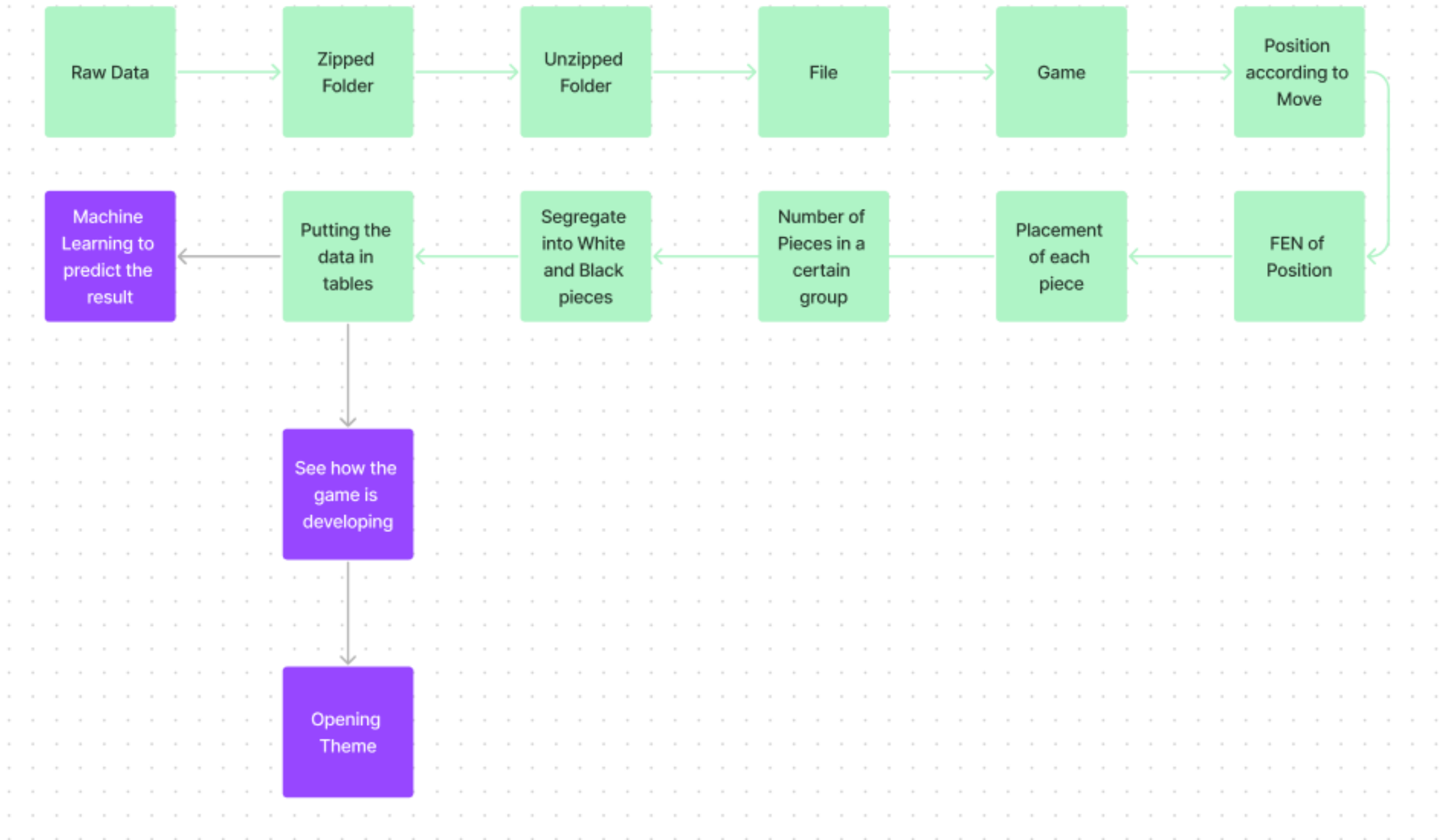

*Figure 2: Various steps involved in processing the data. Firstly, all the raw data was downloaded. Then, it was put into an accessible format by unzipping the folder containing the data. Following this, 960 .pgn files were obtained for each of the 960 starting configurations, where each .pgn file contained 500 games (.pgn is the format in which chess games are stored). Next, the data on each of the games was segregated from each file. This included calculating the number of white and black pieces in each of the regions by getting the .fen file for the specific move required (.fen is the file that tells the placement of each piece on a specific position of a specific game). This data was then into tables to make the process faster. Lastly, specific manipulations were made to the data in the tables to obtain the results, as described in the following sections. All the manipulations and data ordering was done in the python language.*

We organized the chessboard into different regions, such as the Centre, White Kingside, White Queenside, Black Kingside and Black Queenside, as previously mentioned. Each region was associated with specific squares (as mentioned in the Introduction section). This step allowed the algorithm to grasp the spatial arrangement of pieces on the board.

In order to create features for our machine learning model, we extract the .fen from the .pgn files on the specific move number of the particular game that we wanted. A .fen file, short for Forsyth-Edwards Notation, is a compact and standardized notation used to represent a specific position in a chess game. It includes information about the placement of chess pieces on the board, the player to move, castling rights, en passant possibilities, and the number of half-moves and full-moves played so far. Fen files are widely used for sharing and recording chess positions and are easy for both humans and computers to understand and parse. It helps indicate the presence or absence of a piece on a particular square.

We then processed the .fen file for each move, extracting the positions of white and black pieces at that particular moment. The crucial step was comparing the piece positions in the .fen file with the predefined regions and counting the number of white and black pieces in each region. This enabled us to derive the final numerical features for our machine learning model. These features effectively captured the distribution and count of pieces in different parts of the board, providing essential input for our model to analyze and predict outcomes based on the piece positions.

**2.3. Predicting the Outcome**

For all the three models, three separate data sets have been prepared from the raw data to see how accurately the models are able to predict the outcome based on what type of data is fed to them. The following are the three different datasets that were made:

Data Set 1: One game was randomly selected from each of the 960 starting positions and then arrangement of the pieces on move 20 was found, according to the region they were found in. All this data was then represented as a table, which had 960 rows and 11 columns- 5 region columns for White, 5 region columns for Black and 1 for the Result.

Data Set 2: 500 games on each starting position were used to create 960 separate tables, each storing the arrangement of the pieces on move 20. All these 960 different tables were then separately analysed, and the overview of the results have been presented accordingly.

Data Set 3: 500 games on each starting position were used to create 960 separate tables, each storing the arrangement of the pieces on moves 10 through 15. All these 960 different tables were then separately analysed, and the overview of the results have been presented accordingly.

Clearly, Data Set 1 is the smallest data set, followed by Data Set 2 and Data Set 3 is the largest data set that we made after processing the raw data. Further, Data Set 1 the biggest difference between the data sets is that Data Set 1 contains only one game from each starting position, whereas, Data Sets 2 & 3 contain 500 games from each starting position. Additionally, Data Set 2 focuses only on a particular move of each game, whereas Data Set 3 contains 6 moves from each game.

Note: The arrangement of pieces here means finding the number of pieces of a particular color present in a certain region of the chess board.

Selecting move 20 as a rule of thumb for predicting chess outcomes with machine learning is logical due to its midgame positioning. By this point, most opening developments are done, the position stabilizes, and there's enough data for meaningful analysis. The midgame offers strategic depth, with established pawn structures and piece positions, providing valuable context for machine learning models. Around move 20, various moves are possible, creating a challenging prediction task that builds a robust model. Tactical opportunities persist, adding dynamism. Additionally, the midgame's proximity to the endgame is crucial, making it an insightful stage to learn about game outcome factors. A similar set of factors were considered while deciding to analysing the data for moves 10 through 15.

**2.4. Developing an Opening Theme**

All the games that were analyzed in this paper were played by top engines (Rated between ELO Chess Rating ~4000 & ~1800) (*CCRL 40/2 FRC*, 2023), which play even better than humans. This means that all this data contains good quality data which if properly used, can be extremely beneficial for humans to incorporate in their actual games.

For the purpose of this paper, we tried to see how the top engines like to play the game- where they tend to attack in the opening and middlegame. In order to find this out, we used the approach of trying to see the change in position and analyse where the game is shifting. This method was chosen because it was logically sound.

Here, the position of board on Move 1, Move 6, Move 11 and Move 16 (interval of 5 moves) for all the 960 starting positions was collected. Each starting position has 500 different games stored for it, but the each game went on atleast till 16 moves.

This method was carried out primarily in 3 broad steps:

Step 1: The first step is to find the difference of number of pieces in each position between the subsequent moves. The following are the necessary subtractions:

a. Move 6 - Move 1
b. Move 11 - Move 6
c. Move 16 - Move 11

| | Result W A | Result W A | Result W A | Result W A | Result W A |
|---|---|---|---|---|---|
| 2 | 0 | 0 | 1 | -2 | 1 |
| 3 | 0 | 0 | 0 | 0 | 0 |
| 4 | 0 | -2 | 0 | 0 | 2 |
| 5 | 0 | -1 | 0 | 0 | 1 |
| 6 | 0 | -2 | 0 | 0 | 1 |
| 7 | 0 | -3 | 1 | -1 | 3 |
| 8 | 0 | -2 | 0 | -1 | 2 |
| 9 | 0 | -1 | 0 | 0 | 0 |
| 10 | 0 | -1 | 0 | -2 | 2 |
| 11 | 0 | -2 | 0 | 0 | 2 |
| 12 | 0 | -2 | 0 | 0 | 2 |
| 13 | 0 | -1 | 1 | -1 | 1 |
| 14 | 0 | -1 | 0 | 0 | 1 |
| 15 | 0 | -1 | 0 | 0 | 1 |
| 16 | 0 | -2 | 0 | -1 | 3 |
| 17 | 0 | 0 | 0 | -1 | 1 |
| 18 | 0 | -2 | 0 | 0 | 2 |
| 19 | 0 | -4 | 0 | -1 | 4 |
| 20 | 0 | -1 | 0 | 1 | 0 |
| 21 | 0 | -2 | 0 | 1 | 1 |
| 22 | 0 | -2 | 0 | 0 | 1 |
| 23 | 0 | -2 | 0 | -1 | 3 |
| 24 | 0 | -1 | 0 | -1 | 1 |
| | ⋮ | ⋮ | ⋮ | ⋮ | ⋮ |
| 495 | 0 | -2 | 0 | -2 | 4 |
| 496 | 0 | -1 | 0 | 0 | 0 |
| 497 | 0 | -2 | 0 | -1 | 2 |
| 498 | 0 | -1 | 0 | -1 | 2 |
| 499 | 0 | -2 | 1 | -1 | 1 |
| 500 | 0 | -2 | 0 | -1 | 2 |
| 501 | 0 | -2 | 1 | -2 | 2 |

*Figure 3: An example data set after performing Step 1*

Step 2: Once the subtractions are found, we add up all the different differences in a column to find out where the pieces are moving- whether they are going out of a particular region (negative change) or they are coming in a region (positive change)

| (WHITE MOVE 6- WHITE MOVE 1) | | | | |
|---|---|---|---|---|
| Black Queenside | White Queenside | Black Kingsideside | White Kingside | Centre |
| 17 | -834 | 50 | -366 | 810 |

*Figure 4: An example data set after performing Step 2*

Step 3: We are mostly concerned with only the maximum positive change, which will actually give towards which region the position is actually going. Then based on this comparison, we assign them the region towards which the position is changing

| A | B | C | D | E | F | G |
|---|---|---|---|---|---|---|
| White 1 | White 2 | White 3 | Black 1 | Black 2 | Black 3 | Game Code |
| Centre | Centre | Black Q Sic | Centre | Centre | White K Si | BBNNQRKR |

*Figure 5: An example data set after performing Step 3*

## 2. Modelling

We have used three types of machine learning models:

1) KNN Clustering
2) Random Forest
3) Gradient Boosted Trees

In all the models, only discrete values are used instead of continuous values. Although prediction using continuous values might be more accurate, it is illogical to use them as they would only show the potential evaluation of position instead of an outcome. Chess has only three distinct outcomes- White Wins, Draw or Black Wins. For the purpose of this research paper, we have assigned White Wins the value of 1, Draw a 0.5 and Black Wins as 0.

### 3.1. KNN Clustering

The K-nearest neighbors (KNN) clustering model is a versatile machine learning algorithm that can be used for both classification and clustering tasks. While KNN is primarily known as a classification algorithm, it can also be employed as a clustering technique by assigning data points to their nearest neighbours (Tibshirani et al., 2023).

In KNN clustering, the algorithm works as follows. First, a value is chosen for K, representing the number of nearest neighbours to consider. Then, the algorithm calculates the distances between each data point and all other data points in the dataset using a chosen distance metric such as Euclidean or Manhattan distance. The K nearest neighbors are determined for each data point, and based on the majority of neighbours, the point is assigned to a cluster.

This process is repeated for all data points until convergence or a stopping criterion is met. The resulting clusters can then be evaluated using metrics like silhouette coefficient or within-cluster sum of squares. Adjustments to the value of K or the choice of distance metric can be made to refine the clustering.

KNN clustering is relatively simple and easy to understand. In summary, KNN clustering is a non-parametric algorithm that assigns data points to clusters based on the majority of their nearest neighbours. It is flexible and can be used for clustering tasks in addition to classification. By the general rule of thumb, we have chosen K as 31 for Data Set 1 and K as 23 for Data Sets 2 & 3.

### 3.2. Random Forest

Random Forest is a popular machine learning algorithm that combines the power of multiple decision trees to make accurate predictions and classifications. It is known for its versatility, robustness, and ability to handle complex data sets (Tibshirani et al., 2023). The algorithm works by creating an ensemble of decision trees, where each tree is trained on a random subset of the original data. During training, each tree learns to make predictions based on a subset of features randomly chosen at each node. This process introduces diversity among the trees and helps to reduce overfitting.

To make a prediction, the algorithm takes input data and passes it through each individual tree in the forest. Each tree independently predicts the outcome, and the final prediction is determined by majority voting (in classification) or averaging (in regression) the results of all the trees.

Random Forest has several advantages. It can handle large datasets with high dimensionality and a mix of continuous and categorical features. It is also robust against outliers and noisy data.

### 3.3. Gradient Boosted Trees

Gradient Boosted Trees (GBTs) is a machine learning algorithm that combines the power of gradient boosting and decision trees to create a highly accurate predictive model. It is a popular ensemble method used for both regression and classification tasks. GBTs iteratively build an ensemble of weak decision trees, where each subsequent tree is constructed to correct the errors made by the previous trees in the ensemble (Tibshirani et al., 2023). The process of building a GBT involves several steps. Initially, a single decision tree is trained on the dataset. The errors or residuals from this tree are then calculated, and a new decision tree is trained to predict these residuals. The predictions from all the trees in the ensemble are added together to produce the final prediction. This process is repeated iteratively, with each new tree focusing on minimizing the errors made by the previous trees. The learning rate, which determines the contribution of each tree to the ensemble, is a crucial hyperparameter in GBTs.

GBTs have several advantages that contribute to their popularity. Firstly, they can capture complex nonlinear relationships between features and the target variable. Secondly, they handle a variety of data types, including numerical and categorical features. Additionally, GBTs are robust to outliers and can handle missing data effectively. They also provide feature importance measures, enabling better understanding of the importance of different features in the prediction process.

## 4. Result

### 4.1. Outcome Prediction

The following was the data obtained after applying the different machine learning models as shown in Table 2. These are the same Data Sets that were described in the methodology. For Data Set 1, only one outcome is obtained and thus, we have only calculated the mean for it. On the other hand, for Data Sets 2 & 3, the mean, median and maximum observations have been recorded. This is because Data Sets 2 & 3 providing a list of 960 different results, each describing the accuracy for a particular starting configuration. Thus, to summarise that result, the mean accuracy, median accuracy and maximum accuracy features of the data are presented below.

*Table 2: Accuracy table representing the accuracy of various machine learning models across different data sets*

| **Accuracy Table** | | | | |
|---|---|---|---|---|
| **Data Set** | **Parameter** | **KNN** | **Random Forest** | **Gradient Trees** |
| Data Set 1 | Mean | 0.396 | 0.485 | 0.401 |
| Data Set 2 | Median | 0.400 | 0.390 | 0.390 |
| | Mean | 0.398 | 0.391 | 0.394 |
| | Maximum | 0.560 | 0.560 | 0.570 |
| Data Set 3 | Median | 0.390 | 0.375 | 0.380 |
| | Mean | 0.387 | 0.374 | 0.385 |
| | Maximum | 0.540 | 0.570 | 0.570 |

**4.2. Theme Analysis**

After performing the necessary manipulations to the data, we were able to obtain how the top engines like to play in each starting position. The broad data has been recorded in the table below. A more detailed data, so as to decide how to play in the opening/middlegame in each particular starting position, is provided in the Appendix.

*Table 3: Theme Analysis of the starting positions. "Development of White" refers to the side White tends to develop in. Likewise for Black. They are useful markers in telling the direction in which the game is headed.*

| S.No. | Development of White | Development of Black | No. of Starting Positions |
|---|---|---|---|
| 1. | Centre | Centre | 234 |
| 2. | Centre | White Kingside | 27 |
| 3. | Centre | White Queenside | 40 |
| 4. | Black Kingside | Centre | 113 |
| 5. | Black Kingside | White Kingside | 117 |
| 6. | Black Kingside | White Queenside | 79 |
| 7. | Black Queenside | Centre | 148 |
| 8. | Black Queenside | White Kingside | 71 |
| 9. | Black Queenside | White Queenside | 131 |

**5. Discussion**

In this paper, we aimed to predict the result of a chess960 game using different machine learning models and feeding them different types of data sets. The other aim of this paper was to perform a theme analysis of the starting positions.

**5.1. Discussing the Accuracy of the Predictions**

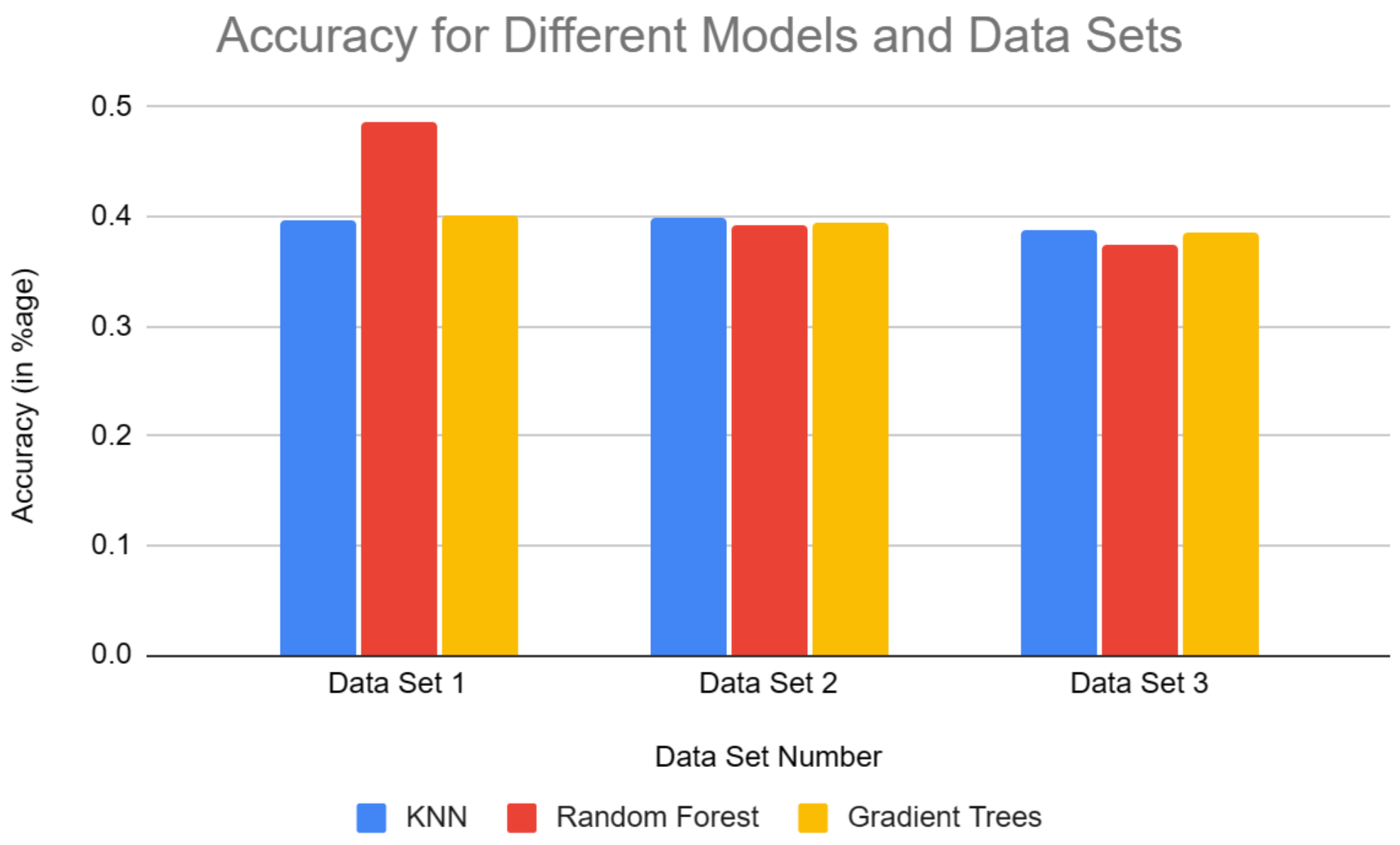

*Figure 6: Comparing the Mean values of different models and data sets*

The results of the machine learning models were quite surprising yet also logical. All the predictions were only ~40% accurate, which is only slightly better than randomly guessing an outcome and getting it right, the probability of which is 33.33% (as there are only 3 outcomes possible). Yet, what was interesting is the model behaved when it had more data to predict the result and also how certain starting positions always had better prediction rates (>50%).

Note: Since in this model we consider 3 outcomes - win, draw or loss. Hence, the chance of randomly guessing the outcome of a game is 33.33%. Hence any value of a model greater than 33.33% is a better than a 3-sided coin toss.

### 5.1.1. Comparing the Different Data Sets

All the three data sets yielded similar results, with Data Set 1 giving marginally better results than the other two Data Sets.

### 5.1.2. Comparing the Different Machine Learning Models

All the three data models have little difference between their predictions, except for in Data Set 1, where the Random Forest model has clearly outperformed all the other models across all the data sets.

### 5.1.3. Disparity in the Accuracies of Certain Starting Positions

An interesting observation was that certain models were able to predict results of certain starting positions much better than those of other starting positions. For example, the maximum accuracy of KNN model on a particular starting position of Data Set 2 was 56% while on another position of the same set was as low as 27%- worse than randomly guessing the outcome.

### 5.1.4. Possible Reasons for Low Accuracy Rate

There could be several reasons for the low accuracy rate. To name a few, they could- overfitting the data, not having enough quality/relevant data, having too many outliers, not having enough data etc. Since the games analyzed have been played by strong chess engines and not actual humans, certain human factors such as time pressure and psychological pressure can be safely eliminated. However, the behavior of chess engines can too vary- sometimes quite differently than that of humans- and that hasn’t been accounted for in this research.

Further, the results of the games in the data sets analysed are somewhat skewed, generally having a lesser probability of the Black side winning overall. This could have also led to some disparity.

Chess960 introduces a greater degree of complexity due to the randomized starting positions. The unfamiliarity of these positions can make it more challenging to predict game outcomes accurately, especially if the model hasn't been extensively trained on Chess960 games.

## 5.2. Discussing the Theme Analysis and Its Importance

When analyzed, it was noticed that in almost all the games (~>90%), both White and Black always went on to develop in the centre till the first 10 moves. Thus, we can safely assume that in most of the cases, it is best to develop in the centre in the early game, till about the first 10 moves. This is also consistent with the general opening theory of chess. The direction where the game is headed is decided between moves 10 to 15. There is a significant number of different strategies that can be used, as demonstrated in the results of the theme analysis. One particularly important thing to notice that both White and Black either always start attacking the other player or develop on the Centre. Overall, in none of the starting positions do either of the players go back to their territory- this is quite logical as one would always like to further expand in chess. This further verifies that the analysis has been done correctly.

The Theme Analysis of the Chess 960 positions can provide valuable insight into how to go about developing a simplified opening theory for each of the starting positions. It also provides some basis for the new players to develop an efficient opening strategy for themselves. However, situations may vary in a game of chess, and there could be many strategies to play a particular position.

## 6. Conclusion

In conclusion, the accuracy of the predictions made by the machine learning models for Chess960 games was found to be only around 40%, slightly better than random guessing. However, certain starting configurations were better, yielding accuracy rates even higher than 50%. Out of all the Data Sets, Data Set 1 performed marginally better than the other two data sets. Moreover, all the three machine learning models also resulted in a similar performance. Next, the results obtained from the Theme Analysis can also prove to be very valuable as it highlights what could be a potential play for the players in every starting configuration.

Overall, this research sheds light on the challenges of predicting Chess960 game outcomes using machine learning and highlights the potential benefits of theme analysis in understanding opening strategies in this variant of chess. Further studies and improvements in data collection and model training may enhance the accuracy of predictions and deepen our understanding of Chess960 gameplay.

## 7. Future Developments & Limitations

There were certain innate limitations to this study. First of all, some of the data may have been overfitted. Secondly, certain conditions and essential parameters were not taken into account, these include- the behavior of the chess engine to different situations, annotations by other players, evaluation of the position, etc. Some of these parameters could be extremely helpful in increasing the accuracy of the model. Lastly, there are class imbalances in the data which may reduce the prediction.

More features can be used for building the model by interviewing chess experts and grandmasters. These kind of parameters can help us design a more efficient algorithm.

## 8. References

[1] Acher, M., & Esnault, F. (2016, April 28). *Large-scale Analysis of Chess Games with Chess Engines: A Preliminary Report*. Retrieved from arxiv.org: https://doi.org/10.48550/arXiv.1607.04186

[2] Friedel, F. (2018, 2 28). *The problem with Chess960*. Retrieved from Chess News: https://en.chessbase.com/post/the-problem-with-chess960

[3] Kumar, N., & Bhayaa, S. (2021). Can Chess ever be Solved. *Turkish Journal of Computer and Mathematics Education*, 1-6.

[4] Tibshirani, R., Hastie, T., James, G., Taylor, J., & Witten, D. (2023). *An Introduction to Statistical Learning.* Springer.

[5] *What is Chess 960 or Fischer Random Chess?* (2021, August 4). Retrieved from Chess.com: https://support.chess.com/article/347-chess960-fischer-random-chess

[6] *CCRL 40/2 FRC - Index*. (2023, August 4). CCRL Blitz Rating List. Retrieved August 10, 2023, from http://www.computerchess.org.uk/ccrl/404FRC/index.html

[7] Gligoric, S. (2003). *Shall We Play Fischerandom Chess?* Pavilion Books.

[8] Sterren, P. v. d. (2009). *Fundamental Chess Openings*. Gambit Publications, Limited.

**9. Appendix**

This section traces the opening theme of each individual starting position. The heading provided is in the form X_Y, where X represents the region in which White pieces tend to develop, and Y represents the region in which Black pieces develop. This is very useful in individually finding out how each specific starting position should be played. For example, the position BBNNQRKR should be played by the players such that White attacks the Black Queenside whereas Black attacks the White Kingside.

The following are the data points:

Black Q Side_White K Side | BBNNQRKR, BBNNRQKR, BBRNKNRQ, BBRNKRQN, BNRNQBKR, BQNNRBKR, BQRKNNRB, BRKBQNNR, BRKQNBNR, BRNBNKRQ, BRNKQBRN, NBQNBRKR, NBQRBNKR, NQNRBBKR, NQNRKBBR, NQRBBKRN, NQRKBBNR, NQRKBBRN, NRBBNQKR, NRBBQNKR, NRBQKNRB, NRKBBNRQ, NRKQNBBR, NRNBBKRQ, NRNKBBQR, NRNKRQBB, NRNQBBKR, NRQBKNBR, NRQKBBRN, NRQKBNRB, NRQKNRBB, NRQKRNBB, QBNRBNKR, QBRNBKRN, QNBNRBKR, QNNBBRKR, QRKNBRNB, RBKNBNRQ, RBKNBRNQ, RBKNNQBR, RBKQBNNR, RBNNKRBQ, RBNQBKNR, RBQKBNRN, RBQKNRBN, RKBBRNNQ, RKBNNBQR, RKBNRBNQ, RKBRNNQB, RKNBBNQR, RKNBQNBR, RKNNBBQR, RKNNRQBB, RKNQBBRN, RKQBBNRN, RKQNBBNR, RKRBBNQN, RKRBNNBQ, RNBBKNQR, RNBBKNRQ, RNBKQBRN, RNBQNKRB, RNKBBNQR, RNKBBQNR, RNKNBBRQ, RNKNRQBB, RNKQBBRN, RNNBBQKR, RNNBKQBR, RNQNKRBB, RQKRNNBB

Black K Side_Centre | BBNNRKQR, BBNNRKRQ, BBNRKQRN, BBNRKRNQ, BBNRKRQN, BBNRNQKR, BBNRQKRN, BBQRNKNR, BBRKNRNQ, BBRNKNQR, BBRNKQRN, BBRQKNNR, BBRQNKNR, BBRQNNKR, BNNQRKRB, BNNRKBRQ, BNQBNRKR, BNQBRNKR, BNQNRBKR, BNRBKNRQ, BNRBNKQR, BNRBNKRQ, BNRKRNQB, BNRQKNRB, BQRKNBRN, BRKBNQRN, BRKNNBRQ, BRKNRBNQ, BRKRNQNB, BRKRQBNN, BRNNKBRQ, BRNNKQRB, BRQBKNNR, NBBQRNKR, NBBRKQRN, NBBRNKQR, NBNRKRBQ, NBRKBNQR, NBRKBQNR, NBRKRNBQ, NBRKRQBN, NBRNBQKR, NBRNQKBR, NNQBBRKR, NNQRKBBR, NQBBRKNR, NQNRBKRB, NQNRKRBB, NQRBKNBR, NQRKBNRB, NQRKRNBB, NQRNKRBB, NRBBKQRN, NRBKNBQR, NRBKQNRB, NRBKRNQB, NRBNKBRQ, NRKBQNBR, NRKQNRBB, NRKQRNBB, NRKRBBNQ, NRKRQNBB, NRQNKBBR, QBBRKNNR, QBNRBKRN, QBRKBNRN, QNBBRNKR, QNBRKBNR, QNBRKBRN, QNBRKNRB, QNBRNBKR, QNRBKNBR, QNRKBBRN, QRBNKBRN, QRKNNBBR, QRNKRBBN, RBBKNQNR, RBBKQNRN, RBBKRNQN, RBBNKQRN, RBBNKRNQ, RBBNQNKR, RBKNBQRN, RBKQNNBR, RBNKBRNQ, RBNKBRQN, RBNKNRBQ, RKBBNQNR, RKBBRNQN, RKBNNRQB, RKBQRNNB, RKBRNQNB, RKNBBQRN, RKNBBRNQ, RKNBNQBR, RKNQBRNB, RKNRBQNB, RKNRNBBQ, RNBKNQRB, RNBKRQNB, RNBNKBRQ, RNBNKRQB, RNKNBQRB,

RNKQBRNB, RNQKBBRN, RNQNBKRB, RQBKNBRN, RQBNKBNR, RQKNBNRB, RQKNNRBB, RQKNRNBB, RQNKNBBR, RQNNBKRB

Centre_Centre | BBNQNRKR, BBQRKNNR, BBQRKRNN, BBQRNKRN, BBRKNQRN, BBRKQRNN, BBRKRQNN, BBRQKRNN, BNNRKQRB, BNNRQKRB, BNQBRKNR, BNQBRKRN, BNQRKBNR, BNQRKBRN, BNQRKRNB, BNQRNKRB, BNRBKNQR, BNRBKQRN, BNRBKRNQ, BNRBKRQN, BNRBQKNR, BNRBQKRN, BNRKNQRB, BNRKNRQB, BNRKQBNR, BNRKQBRN, BNRKQRNB, BNRKRBNQ, BNRKRBQN, BNRKRQNB, BNRNKQRB, BNRNKRQB, BNRNQKRB, BNRQKBRN, BNRQKRNB, BNRQNBKR, BQNBRKNR, BQNRKRNB, BQNRNKRB, BQRBKNNR, BQRBKRNN, BQRBNKNR, BQRBNKRN, BQRKNRNB, BQRKRBNN, BQRNKBNR, BQRNKNRB, BQRNKRNB, BQRNNBKR, BQRNNKRB, BRKBNNQR, BRKBNQNR, BRKBNRQN, BRKBQRNN, BRKBRQNN, BRKNNQRB, BRKNNRQB, BRKNQNRB, BRKNRBQN, BRKNRQNB, BRKQNBRN, BRKQRBNN, BRNKNBQR, BRNKNQRB, BRNKNRQB, BRNKQRNB, BRNQNKRB, BRQBKRNN, BRQBNKNR, BRQBNKRN, BRQKNBNR, BRQKNRNB, BRQNKBNR, BRQNKNRB, BRQNKRNB, BRQNNKRB, NBBNRKQR, NBBQNRKR, NBBQRKNR, NBBQRKRN, NBBRKQNR, NBBRKRNQ, NBBRKRQN, NBBRNKRQ, NBBRQKNR, NBBRQKRN, NBBRQNKR, NBNQRKBR, NBQNRKBR, NBQRBKNR, NBQRKNBR, NBQRKRBN, NBRKBRNQ, NBRQBKNR, NBRQBKRN, NBRQKNBR, NNBQRBKR, NNBQRKRB, NNBRKBQR, NNBRKBRQ, NNBRKRQB, NNBRQKRB, NNQRBKRB, NNQRKRBB, NNRBBKRQ, NNRBBQKR, NNRBKRBQ, NNRBQKBR, NNRKBBQR, NNRKBBRQ, NNRKBQRB, NNRKBRQB, NNRKQBBR, NNRKQRBB, NNRKRBBQ, NNRKRQBB, NNRQBKRB, NNRQKBBR, NNRQKRBB, NQBNRKRB, NQBRKNRB, NQRBBKNR, NQRKBRNB, NQRNBKRB, NRBKNRQB, NRBKRBQN, NRBNKBQR, NRBNKRQB, NRBNQKRB, NRKBBRNQ, NRKBRNBQ, NRKBRQBN, NRKNBBQR, NRKNBBRQ, NRKNBQRB, NRKNBRQB, NRKNQBBR, NRKNQRBB, NRKNRBBQ, NRKNRQBB, NRKQRBBN, NRKRNBBQ, NRQBBKNR, NRQKRBBN, NRQNBKRB, NRQNKRBB, QBBNRKNR, QBBNRKRN, QBBNRNKR, QBBRKNRN, QBBRKRNN, QBBRNKRN, QBBRNNKR, QBNRBKNR, QBNRKNBR, QBNRKRBN, QBNRNKBR, QBRKBNNR, QBRKBRNN, QBRNKRBN, QNBBRKNR, QNBNRKRB, QNBRKRNB, QNNBRKBR, QNNRBBKR, QNRBBKNR, QNRBKRBN, QNRKBRNB, QNRKRBBN, QNRKRNBB, QRBBKRNN, QRBBNKNR, QRBBNKRN, QRBKRBNN, QRBNKBNR, QRKBNRBN, QRKNBBNR, QRKNNRBB, QRKNRNBB, QRNBKRBN, RBBKNQRN, RBBKQNNR, RBBKQRNN, RBBNNQKR, RBBQKNNR, RBBQKRNN, RBBQNKRN, RBKNBRQN, RBKNNRBQ, RBKNQNBR, RBKNQRBN, RBKNRNBQ, RBKQNRBN, RBKRNNBQ, RBNQNKBR, RBQKBRNN, RKBBNRNQ, RKBBRQNN, RKBNQRNB, RKBNRQNB, RKBQNBNR, RKBRQBNN, RKNBRQBN, RKNNBRQB, RKQNNBBR, RKRNBQNB, RNBBKQNR, RNBBKRNQ, RNBBNKRQ, RNBBQKNR, RNBKNBRQ, RNBKNRQB, RNBKRBQN, RNBNQKRB, RNBQKRNB, RNKBBRNQ, RNKBQRBN, RNKNBRQB, RNKNQRBB, RNKQRBBN, RNKRBBNQ, RNKRBQNB, RNKRQBBN, RNNKBRQB, RNNKQRBB, RNNQBBKR, RQBBKRNN, RQBBNKRN, RQBKRBNN, RQBNKRNB, RQKBNRBN, RQNBNKBR, RQNKBRNB, RQNKNRBB

Centre_White K Side | BBNQRKNR, BBNRNKQR, BBNRQKNR, BNRBKQNR, BNRKQNRB, BQNRKNRB, BQRNKBRN, BRKNRNQB, NQBNRBKR, NRBBQKNR, NRBKQBNR, NRBQNKRB, NRNKRBBQ, QBRNKNBR, QNBRNKRB, QNNRKRBB, QRNNBKRB, RBBKNNRQ, RBKNRQBN, RBNQBKRN, RBQNBKNR, RKBBQNNR, RKQNBNRB, RNBNKQRB, RNKBNQBR, RNKRNQBB, RNNKBQRB

Black Q Side_Centre | BBNQRKRN, BBNRNKRQ, BBRKNRQN, BBRNNKQR, BBRQNKRN, BNQRNBKR, BNRQNKRB, BQNRKBNR, BQNRNBKR, BQRBNNKR, BQRKNBNR, BRKBNRNQ,

BRKBRNNQ, BRKBRNQN, BRKNQRNB, BRKQNRNB, BRKRNBQN, BRNBKRQN, BRNBQKRN, BRNKQBNR, BRNKRQNB, BRNQKBNR, BRNQKRNB, BRQBNNKR, BRQKNBRN, BRQKNNRB, BRQNNBKR, NBBRKNRQ, NBNRQKBR, NBQRNKBR, NBRKQNBR, NBRKQRBN, NBRNBKQR, NBRQKRBN, NNBRKQRB, NNBRQBKR, NNQBRKBR, NNQRBBKR, NNRBBKQR, NQBBRKRN, NQBRKBNR, NQBRKBRN, NQBRKRNB, NQBRNKRB, NQNBBRKR, NQNBRKBR, NQRBKRBN, NQRBNKBR, NQRKNRBB, NRBBKQNR, NRBBNKQR, NRBKNQRB, NRBKQRNB, NRBKRBNQ, NRBKRQNB, NRBNKQRB, NRBNQBKR, NRBQKBRN, NRBQKRNB, NRBQNBKR, NRKBBQNR, NRKQBRNB, NRNBKRBQ, NRNQBKRB, NRNQKRBB, NRQBKRBN, NRQBNKBR, NRQKBRNB, NRQKNBBR, QBNNRKBR, QBRKNNBR, QBRKNRBN, QBRNBNKR, QBRNNKBR, QNRBBKRN, QNRKBBNR, QNRKNBBR, QNRKNRBB, QNRNBBKR, QRBBKNNR, QRBKNBRN, QRBKNNRB, QRBKNRNB, QRBNKRNB, QRKBBRNN, QRKRNNBB, QRNBBKRN, QRNBNKBR, QRNKBRNB, QRNKNBBR, RBBKRNNQ, RBBKRQNN, RBBNKNRQ, RBBNKQNR, RBBNQKNR, RBBQNNKR, RBKQBRNN, RBKRNQBN, RBKRQNBN, RBNKBQNR, RBNKBQRN, RBNKNQBR, RBNKQNBR, RBNKQRBN, RBNKRNBQ, RBNKRQBN, RBNNBKQR, RBNQKRBN, RBQNNKBR, RKBNNQRB, RKBQNBRN, RKBRNBNQ, RKNBRNBQ, RKNNRBBQ, RKNQNRBB, RKNQRBBN, RKNRBBNQ, RKNRQBBN, RKQNRNBB, RNBBKRQN, RNBKQBNR, RNBKRBNQ, RNBQKBNR, RNBQKBRN, RNBQKNRB, RNBQNBKR, RNKQNRBB, RNKQRNBB, RNKRBBQN, RNKRQNBB, RNNKBBQR, RNNKQBBR, RNNKRBBQ, RNNKRQBB, RNQBBKRN, RNQKBNRB, RNQKBRNB, RNQKRBBN, RNQKRNBB, RNQNBBKR, RNQNKBBR, RQBBKNRN, RQBKNRNB, RQKNBRNB, RQKNRBBN, RQNBKRBN, RQNKBBRN, RQNKRBBN

Black K Side_White K Side | BBNQRNKR, BBNRKNQR, BBNRKNRQ, BBNRQNKR, BBQNNRKR, BBQNRKRN, BBQNRNKR, BBQRKNRN, BBRKNNQR, BBRKNNRQ, BBRKNQNR, BBRKQNNR, BBRKQNRN, BBRKRNNQ, BBRNKQNR, BBRNKRNQ, BBRNNKRQ, BBRNNQKR, BBRNQKNR, BBRNQNKR, BNNBRQKR, BNNRKBQR, BNNRKRQB, BNNRQBKR, BNQNRKRB, BNRBNQKR, BNRKNBQR, BNRNKBRQ, BRKBQNRN, BRKNNBQR, BRKNQBNR, BRKRQNNB, BRNKQNRB, BRNKRBQN, BRNKRNQB, BRQBKNRN, NBBNQRKR, NBBNRKRQ, NBBNRQKR, NBBRKNQR, NBNQBRKR, NBNRBKQR, NBNRBKRQ, NBNRBQKR, NBRKBNRQ, NBRKBQRN, NBRKBRQN, NBRKNQBR, NBRKNRBQ, NBRNBKRQ, NBRNKQBR, NBRQBNKR, NNBBRKRQ, NNBBRQKR, NNRBKQBR, NQBBRNKR, NQRBBNKR, NRBBKNQR, NRBBKNRQ, NRBBKRNQ, NRBBKRQN, NRBBQKRN, NRBKQBRN, NRKBBNQR, NRKBBQRN, NRKBBRQN, NRKRBNQB, NRNBBQKR, NRNKBBRQ, NRNKBQRB, NRNKQBBR, QBBRNKNR, QNBBNRKR, QNBBRKRN, QNNRBKRB, QNRKBNRB, QRKBNNBR, QRKNBNRB, QRKRBBNN, QRNKBBRN, QRNKBNRB, RBBNKRQN, RBKQBNRN, RBKRBNNQ, RBNKBNQR, RBNKBNRQ, RBNNBKRQ, RBNQBNKR, RKBBNNQR, RKBBNQRN, RKBBQNRN, RKBNNBRQ, RKBNQBNR, RKBNQBRN, RKBNQNRB, RKBNRNQB, RKBQNNRB, RKBQNRNB, RKNBBNRQ, RKNBBQNR, RKNQBBNR, RKNQBNRB, RKQBNNBR, RKQNBBRN, RKRBQNBN, RNBBKQRN, RNBKNBQR, RNBKQNRB, RNBKRNQB, RNBNKBQR, RNKBBQRN, RNKBQNBR, RNKNBBQR, RNKNQBBR, RNKQBBNR, RNNBQKBR, RNNQKBBR

Black K Side_White Q Side | BBNRKQNR, BBQRNNKR, BBRQKNRN, BNNBQRKR, BNNBRKQR, BNNBRKRQ, BNQRKNRB, BNRNKBQR, BQNBRKRN, BQNBRNKR, BQRBKNRN, BRKQNNRB, BRNBKQNR, BRNBQKNR, BRNNQKRB, NBNRKQBR, NBQRBKRN, NBRNKRBQ, NNBBQRKR, NQRNBBKR, NQRNKBBR, NRBBNKRQ, NRKQBBNR, NRKQBNRB, NRKRBBQN, NRKRNQBB, NRNKBRQB, NRNKQRBB, NRNQKBBR, NRQBBNKR, QBBNNRKR, QBRNBKNR, QNRNBKRB, QNRNKBBR, QRBBKNRN, QRBKNBNR, QRKNBBRN, QRNBBNKR, QRNNKRBB, RBBKNNQR, RBBKNRQN, RBBNKNQR, RBBNQKRN, RBBQNKNR, RBKRBQNN, RBNNKQBR, RBQKNNBR,

RBQKRNBN, RKBBNNRQ, RKBBNRQN, RKBBQRNN, RKBNRBQN, RKBRQNNB, RKNNBQRB, RKNNQRBB, RKNQRNBB, RKNRBBQN, RKNRBNQB, RKQBBNNR, RKQBNRBN, RKQRBBNN, RKRBBNNQ, RKRBNQBN, RKRNBBQN, RKRNBNQB, RKRQBBNN, RKRQBNNB, RNBBNQKR, RNBNQBKR, RNKBBNRQ, RNKQBNRB, RNKQNBBR, RNNQBKRB, RNNQKRBB, RQKBBNNR, RQKBNNBR, RQNBBNKR, RQNKBBNR, RQNNKRBB

Centre_White Q Side | BBQNRKNR, BNRKNBRQ, BNRQKBNR, BQNBNRKR, BRKNQBRN, BRKRNBNQ, BRNBNQKR, BRQNKBRN, NBBRNQKR, NBRQNKBR, NNRQBBKR, NQBBNRKR, NRKBQRBN, NRKRBQNB, NRKRQBBN, QBRKRNBN, QNNRKBBR, QNRNKRBB, QRBNKNRB, QRBNNKRB, QRKRNBBN, RBBNNKQR, RBKNBQNR, RBQKBNNR, RKBQRBNN, RKBRNBQN, RKNNBBRQ, RKQNRBBN, RKRNQNBB, RNBKQRNB, RNKBBRQN, RNQBKNBR, RQBBKNNR, RQBBNKNR, RQBBNNKR, RQBNKBRN, RQBNNKRB, RQKBBRNN, RQKRBNNB, RQKRNBBN
Black Q Side_White Q Side | BBRKRNQN, BBRNQKRN, BNNQRBKR, BNRBQNKR, BQNNRKRB, BQNRKBRN, BQRKRNNB, BRKBNNRQ, BRKQRNNB, BRKRNNQB, BRNBKNQR, BRNBKNRQ, BRNBKQRN, BRNBKRNQ, BRNBNKQR, BRNBQNKR, BRNKNBRQ, BRNKRBNQ, BRNNKBQR, BRNNKRQB, BRNNQBKR, BRNQKBRN, BRNQKNRB, BRNQNBKR, BRQKRBNN, BRQKRNNB, NNBBRKQR, NQBRNBKR, NQRKNBBR, NQRKRBBN, NRBKNBRQ, NRBQKBNR, NRKBNQBR, NRKBNRBQ, NRKQBBRN, NRNBBKQR, NRNBKQBR, NRNBQKBR, NRQBBKRN, NRQKBBNR, NRQNBBKR, QBNNBRKR, QNRBBNKR, QNRBNKBR, QRBBNNKR, QRBKRNNB, QRBNNBKR, QRKBBNNR, QRKBBNRN, QRKBRNBN, QRKNRBBN, QRKRBNNB, QRNBBKNR, QRNBKNBR, QRNKBBNR, QRNKNRBB, QRNKRNBB, QRNNBBKR, QRNNKBBR, RBBKNRNQ, RBBNNKRQ, RBBQKNRN, RBKNBNQR, RBKQRNBN, RBKRBNQN, RBNNBQKR, RBNNQKBR, RBNQKNBR, RBQNBKRN, RBQNBNKR, RBQNKNBR, RBQNKRBN, RKNBBRQN, RKNBNRBQ, RKNBQRBN, RKNNQBBR, RKNQNBBR, RKNRNQBB, RKNRQNBB, RKQBBRNN, RKQBRNBN, RKQNBRNB, RKQNNRBB, RKQRBNNB, RKQRNBBN, RKQRNNBB, RKRBBQNN, RKRNBBNQ, RKRNNBBQ, RKRNNQBB, RKRNQBBN, RKRQNBBN, RKRQNNBB, RNBBNKQR, RNBBQKRN, RNBBQNKR, RNKBNRBQ, RNKBRNBQ, RNKBRQBN, RNKNRBBQ, RNKRBNQB, RNKRNBBQ, RNNBBKQR, RNNBBKRQ, RNNBKRBQ, RNNKBBRQ, RNQBBKNR, RNQBBNKR, RNQBKRBN, RNQBNKBR, RNQKBBNR, RNQKNBBR, RNQKNRBB, RQBKNBNR, RQBKNNRB, RQBKRNNB, RQBNKNRB, RQBNNBKR, RQKBBNRN, RQKBRNBN, RQKNBBNR, RQKNBBRN, RQKNNBBR, RQKRBBNN, RQNBBKNR, RQNBBKRN, RQNBKNBR, RQNKBNRB, RQNKRNBB, RQNNBBKR, RQNNKBBR